\documentclass[conference]{IEEEtran}
\IEEEoverridecommandlockouts
\usepackage{cite}
\usepackage{amsmath,amssymb,amsfonts}
\usepackage{algorithmic}
\usepackage{graphicx}
\usepackage{textcomp}
\usepackage{xcolor}
\usepackage{listings}
\usepackage{microtype}
\usepackage{booktabs} 
\usepackage{url}

\definecolor{codegreen}{rgb}{0,0.6,0}
\definecolor{codegray}{rgb}{0.5,0.5,0.5}
\definecolor{codepurple}{rgb}{0.58,0,0.82}
\definecolor{backcolour}{rgb}{1, 1, 1}
\lstdefinestyle{mystyle}{
    backgroundcolor=\color{backcolour},   
    commentstyle=\color{codegreen},
    keywordstyle=\color{magenta},
    numberstyle=\tiny\color{codegray},
    stringstyle=\color{codepurple},
    basicstyle=\ttfamily\footnotesize,
    breakatwhitespace=false,         
    breaklines=true,                 
    captionpos=b,                    
    keepspaces=true,                 
    numbers=left,                    
    numbersep=5pt,                  
    showspaces=false,                
    showstringspaces=false,
    showtabs=false,                  
    tabsize=2
}
\def\BibTeX{{\rm B\kern-.05em{\sc i\kern-.025em b}\kern-.08em
    T\kern-.1667em\lower.7ex\hbox{E}\kern-.125emX}}
\begin{document}

\title{A Scalable and Reproducible System-on-Chip Simulation for Reinforcement Learning}

\author{\IEEEauthorblockN{1\textsuperscript{st} Tegg Taekyong Sung}
\IEEEauthorblockA{
\textit{EpiSys Science, Inc.}\\
Poway, CA \\
tegg@episyscience.com}
\and
\IEEEauthorblockN{2\textsuperscript{nd} Bo Ryu}
\IEEEauthorblockA{
\textit{EpiSys Science,Inc.}\\
Poway, CA \\
boryu@episyscience.com}
}

\maketitle

\begin{abstract}
Deep Reinforcement Learning (DRL) underlies in a simulated environment and optimizes objective goals. By extending the conventional interaction scheme, this paper proffers gym-ds3, a scalable and reproducible open environment tailored for a high-fidelity Domain-Specific System-on-Chip (DSSoC) application. The simulation corroborates to schedule hierarchical jobs onto heterogeneous System-on-Chip (SoC) processors and bridges the system to reinforcement learning research. We systematically analyze the representative SoC simulator and discuss the primary challenging aspects that the system (1) continuously generates indefinite jobs at a rapid injection rate, (2) optimizes complex objectives, and (3) operates in steady-state scheduling. We provide exemplary snippets and experimentally demonstrate the run-time performances on different schedulers that successfully mimic results achieved from the standard DS3 framework and real-world embedded systems.
\end{abstract}


\section{Introduction} \label{introduction}
Deep reinforcement learning (deep RL or DRL) has breakthrough performances in tactical games~\cite{silver2016mastering,vinyals2019grandmaster,schrittwieser2020mastering} and robotics~\cite{levine2016end,peng2018sim,andrychowicz2020learning}. To these successes, a systematized RL-perspective environment is essential to proceed with sequential interactions. The prior works in environment developments for various domains are robotic manipulation~\cite{todorov2012mujoco,coumans2016pybullet} and vehicle maneuver~\cite{zamora2016extending,shah2018airsim}. The critical mechanism underlying systems is that the agent must continuously interact with the simulation and receive the necessary information straightforwardly~\cite{1606.01540}.

As a universal problem, resource allocation is associated with various problems, including clustering and wireless communication~\cite{rico2010scalable,ousterhout2013sparrow,verma2015large,vega2020stomp}. Despite the successes in various scheduling applications, previous research has overlooked heterogeneous many-core systems. As a representative, Domain-Specific System-on-chip Simulation (DS3) is a high-fidelity discrete-event simulator targetted to the Domain-Specific System-on-Chip (DSSoC) application and faithfully mimics the real-world hardware performance.~\cite{arda2020ds3}.

To facile RL agents' design tailored to real-world systems, this paper introduces a gym-ds3 environment built upon the DS3 framework accessible to the RL research community. We systematically analyze the standard DS3 framework and elaborate fundamental challenging standpoints in designing RL agents to the DS3 framework. Contrast to available scheduling applications, the agent in the DS3 system requires to tackle various joint action sets in complex dynamics due to the hierarchical task dependency and fast job injection rate. Furthermore, we experimentally demonstrate that the scheduling performances operated in the proposed gym-ds3 framework equivalent to the performances in the standard DS3 framework. We publicly release code at \url{https://github.com/EpiSci/gym-ds3}.

\section{Related Work} \label{sec:related}
The most relevant approach with this paper is the Park platform, which is a unified open Gym framework~\cite{1606.01540} for ten types of real-world simulators, including cluster scheduling, video streaming, network congestion, memory caching, and circuit design~\cite{mao2019park}. Park categorizes environments into different challenging problems in systems and provides exemplary RL algorithms.

Concerning scheduling applications, various existing simulations are developed by real-time execution. Formerly, Sparrow builds on a decentralized design that concurrently operates scheduler decisions for cluster jobs~\cite{ousterhout2013sparrow}. Borg manages large-scale clusters and aims to minimize the fault-recovery time in run-time failures for scheduling decisions~\cite{verma2015large}. Bose et al. summarize a secure and resilient embedded SoCs applicable in the power-efficient autonomous vehicles domain~\cite{bosesecure}.

As a representative simulator in the SoC application, the DS3 framework positions a \textit{de facto} benchmarking simulator in active research. DeepSoCS is the first deep RL hybrid scheduler that outperforms run-time performances over standard schedulers provided in the DS3 framework~\cite{sung2020deepsocs}. Krishnakumar et al. apply imitation learning technique to maintain competitive run-time performances and optimize power dissipation and energy efficiency~\cite{krishnakumar2020runtime}. HiLITE is dynamic power management incorporated into the DS3 framework that utilizes imitation learning to optimize energy efficiency~\cite{sartor2020hilite}.

\section{Background} \label{sec:background}
The presented gym-ds3 is built on the Gym framework, which is an open interface connecting from the simulation and reinforcement learning algorithms~\cite{1606.01540}. The system is mainly comprised of \texttt{reset} function to warm up the environment to initialize relevant components and \texttt{step} function to receive an agent action and respond by immediate reward and the next state acquired by the system dynamics. Thereby, the provided action-and-response methodology is suitable for RL agent workflow.

\begin{figure}[!t]
\centering
\includegraphics[height=0.6\linewidth]{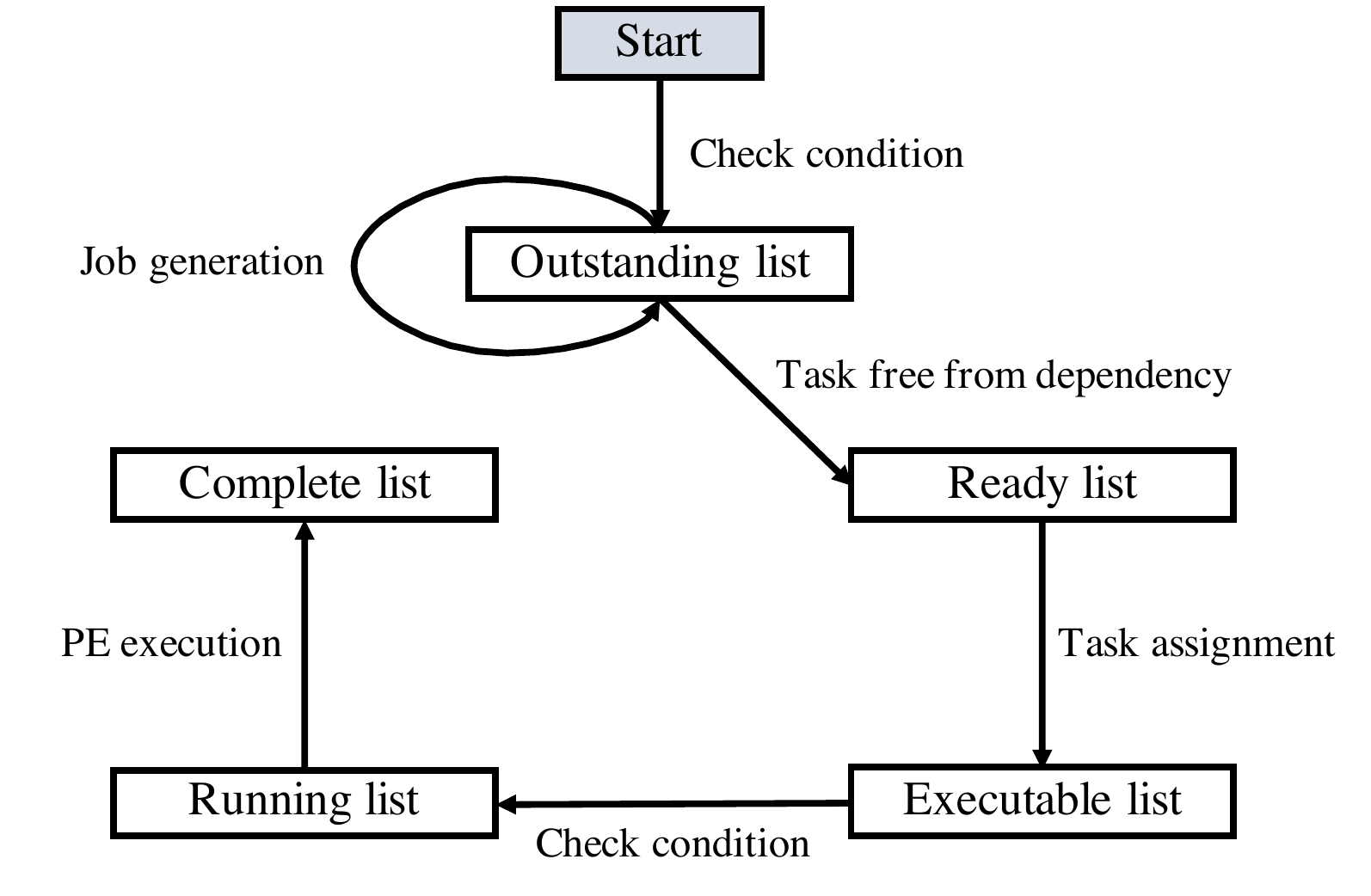}
\caption{A DS3 framework workflow. Here, the system operates with a single processing element.}
\label{fig:ds3_workflow}
\end{figure}

We begin by elaborating the standard DS3 framework specialized in scheduling hierarchical jobs to heterogeneous resources for an SoC application. The system processes in a real-time operation running by flops (floating-point operations per second) and designed with non-preemptive execution. An overall workflow is illustrated in Figure~\ref{fig:ds3_workflow}. The simulation comprises a job generator, distributed processing elements (PEs), scheduling policy, and simulation kernel governing task statuses.

\subsection{Jobs and resources} \label{sec:background:jobs_n_resources}
\begin{figure}[!t]
  \centering
    \includegraphics[height=0.8\linewidth]{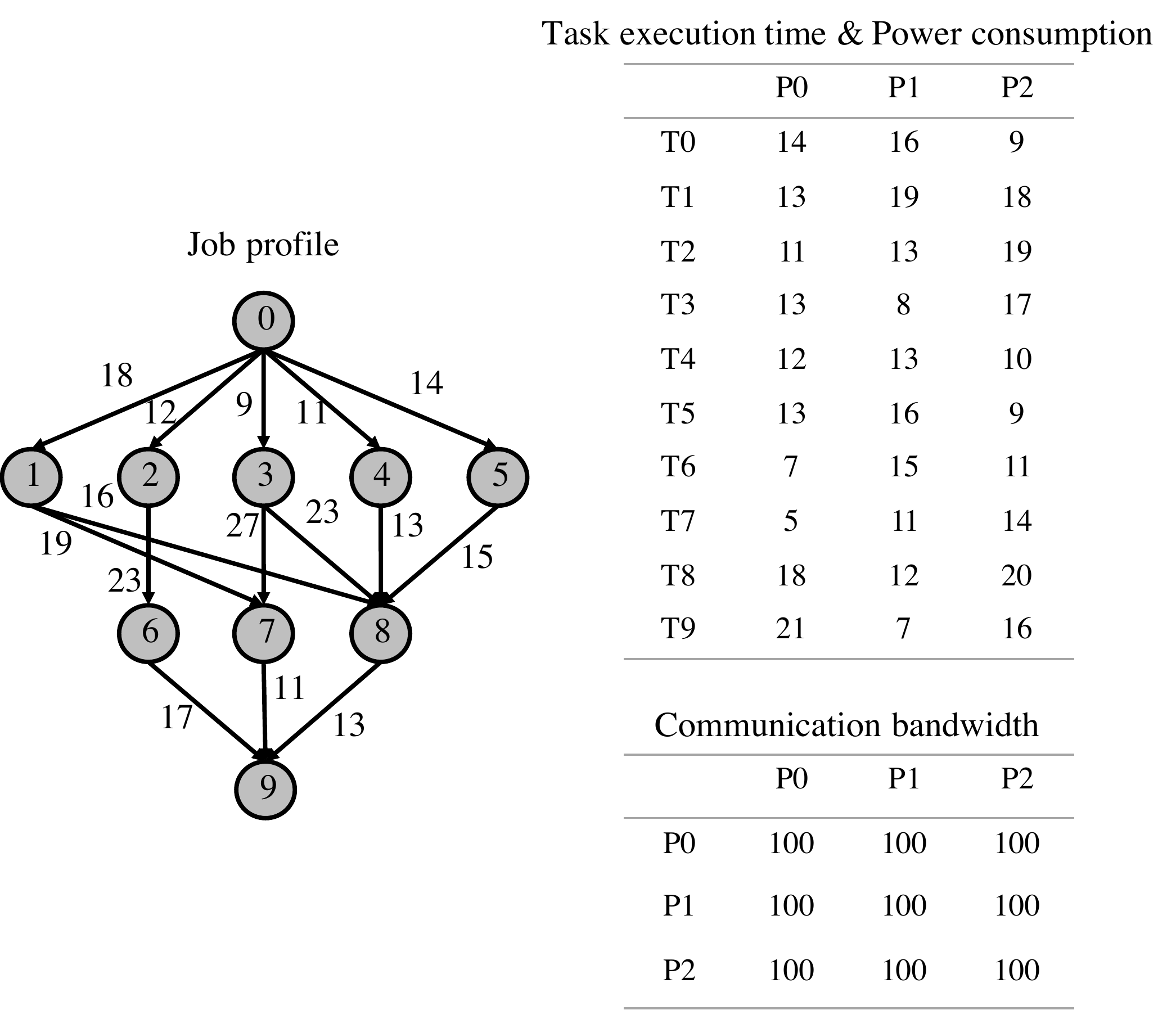}
    \caption{A canonical job and a chart of resource profile~\cite{topcuoglu2002performance}.}
    \label{fig:top_profile}
\end{figure}

The DS3 framework proceeds the kernel using a job (workload application) and resource (processing element) profiles originally targeted to wireless communication and the radar processing domain. As shown in Figure~\ref{fig:top_profile}, a job is depicted by a \textit{Directed Acyclic Graph} (DAG) $G=(V,E)$, where vertices correspond to tasks and edges to communication cost. The node values denote the task number, and the edge values entail data transmission delay accrued to the resource switching. The graph topology represents the dependencies between the tasks. Unlike numerous studies designed for cluster applications that specifying the job duration in the profiles, the SoC workload spanning time is defined by the assigned processing elements' functionalities from the scheduling policy. The resources or processing elements (PEs)\footnote{Resources represent the raw profile information, and processing elements represent essential operating components execute tasks.} are structured with different functionalities in task run-time and energy consumption. Each PE also contains a list of Operating Performance Points (OPP), which characterize the supported frequency and voltage points. Thereby, the scheduling policy directly reflects the overall performance.

\subsection{Job generator} \label{sec:background:job_generator}
One key property in DS3 is that the job generator continuously generates an indefinite number of jobs based on the input profile. The job queue holds $C$ jobs, where $C \leq T$. Here, $C$ denotes the number of jobs and $T$ the length of the job queue. A scale value denotes the frequency of job generation. The smaller scale entails high frequency and reflects the faster injection rate.

Once the job is generated, the system identifies job topology and distributes tasks into separate lists of statuses. As shown in Figure~\ref{fig:ds3_workflow}, the tasks free from dependency load to the `ready list' and those having child dependencies remain in the `outstanding list'. As of multiple jobs awaited in the job queue, tasks with different job instances are arbitrarily overlapped. The more overlapping workloads derive the system more dynamic and stochastic.

\subsection{Scheduling policy} \label{sec:background:scheduler}
The scheduler executes in run-time and essentially allocates each task in the ready list to available PE. The scheduling policy irregularly interacted with the constant clock signal (flops) due to the task dependency and overlapped task traces. After all assignments, the tasks move to the `executable list' and await the PE execution. If the designated PE is idle, the distributed PEs concurrently run the assigned tasks in parallel, and the task transits to the `running list'.  After execution, the task finally moves to the `complete list'.

Assume that the job is composed of $N$ tasks. The individual task run-time is the cumulative sum of PE run-time and data transmission delay, which is calculated with the job edge and PE performance as depicted in Figure~\ref{fig:top_profile}. The task has waiting time when remaining time in the executable list and response time to the cumulative waiting time. The average response time (ART) denotes the rate between task execution time and task waiting time for all tasks. The scheduler operating in small ART generally indicates adequate performances. The scheduling objective is varied by the designing goals, for instance, minimizing average latency\footnote{The latency is proportional to the number of completed jobs and inversely proportional to the cumulative execution time.}, power dissipation, or energy consumption.

\section{Proposed approach} \label{sec:proposed}
The prototypical DS3 framework is developed by SimPy~\cite{matloff2008introduction} that is a process-based discrete-event simulation framework. SimPy built-in simulation can be operated in real-time, and the simulation kernel with multiple instances can be executed in parallel. This notion, however, is difficult to directly match the interaction time step in the reinforcement learning perspective. We thereby derive a representative environment that supports the standard Gym scheme upon the DS3 framework.

The scheduling agent can be designed as a global control or distributed control, depending on the design perspective. For the latter case, independent PE can be referred to multiple agents collaborating to complete tasks quickly. Moreover, gym-ds3 supports Ray, a distributed operating framework manageable to scalable training~\cite{moritz2018ray}. For the current version of gym-ds3, the state indicates the simulation information (i.e., task/job/PE statuses, relevant task time), the action space, a joint set on which task allocates to which PEs. The reward function is stated with the average job duration. Users can modify state, action, and reward statements afterward. For the following sections, we pose the challenging critical standpoints for RL agents in DS3 simulation.

\subsection{Challenges} \label{sec:proposed:challenges}
We highlight the main challenging standpoints arising from designing RL agents in the gym-ds3 environment. First, the agent must tackle a varying number of actions. The action of a scheduling agent is a set of tuples composed of two primitive actions, task and PE selections. Systematically, DS3 fetches more than one specifically designed graph-structured job profile. One straightforwardly considers selecting actions for the tasks in the ready list. In that sense, for various task dependencies on different job profiles, the number of remaining tasks is constantly changing. Furthermore, the ready tasks are arbitrarily overlapped by the multiple injecting jobs. Hence, the action associates with a variable joint action problem. Second, the complexity in action spaces exponentially increases to the number of jobs and heterogeneous resources. The disparate functionalities from the distributed PEs and the combination of tasks in the mixed job topology increase complexity in sequential action selection. Third, the indefinite jobs generated at a fast injection rate cause the system dynamics more complex. One of the main differences between SoC and the clustering domain is that the SoC workloads essentially run in a short duration but much faster workload injection.

\begin{figure}[h]
\centering
\includegraphics[height=.48\textwidth]{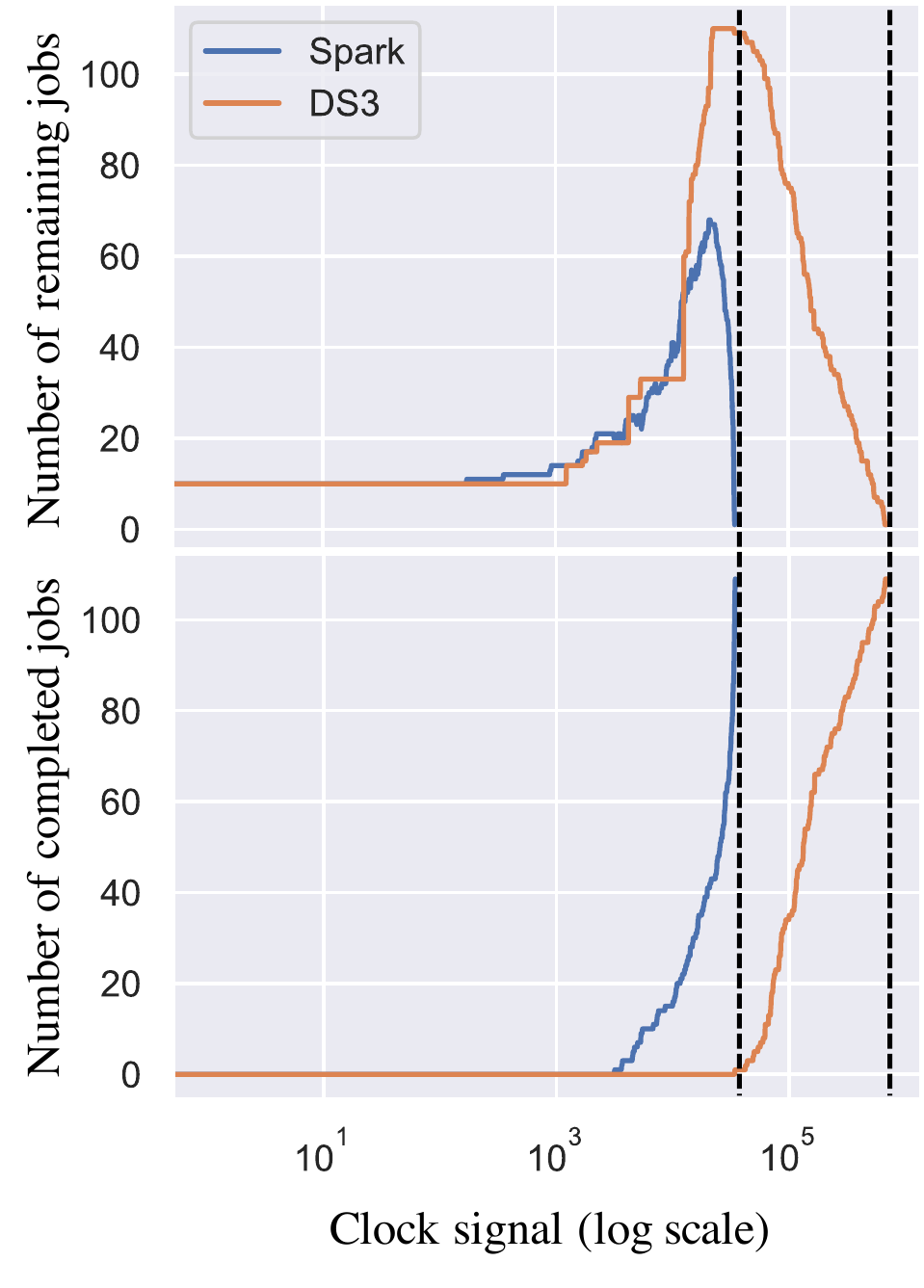}
\caption{An evaluation of the number of remaining and completed jobs for DS3 and Spark simulations.}
\label{fig:rem_comp_jobs}
\end{figure}
Figure~\ref{fig:rem_comp_jobs} depicts the analysis of job rates for DS3 and Spark simulations~\cite{mao2019learning}. Here, we modify simulations with the same range of running flops and the number of injected jobs. We evaluate scheduling performance with heuristic schedulers. Due to innate development, DS3 completes jobs faster than Spark by a factor of 10.

\subsection{Steady-state scheduling} \label{sec:proposed:steady_state}
Absolute makespan minimization of scheduling heterogeneous resource is NP-hard in most practical situations~\cite{ausiello2012complexity,shirazi1995scheduling}. The steady-state scheduling circumvents this difficulty by considering asymptotic optimality~\cite{beaumont2005steady,bertsimas1999asymptotically}. In DS3 simulation, jobs are indefinitely generated, and the scheduling performance is evaluated starting from the steady-state that the jobs are fully stacked to the job queue. 

\begin{figure}[h]
    \centering
    \includegraphics[height=0.5\linewidth]{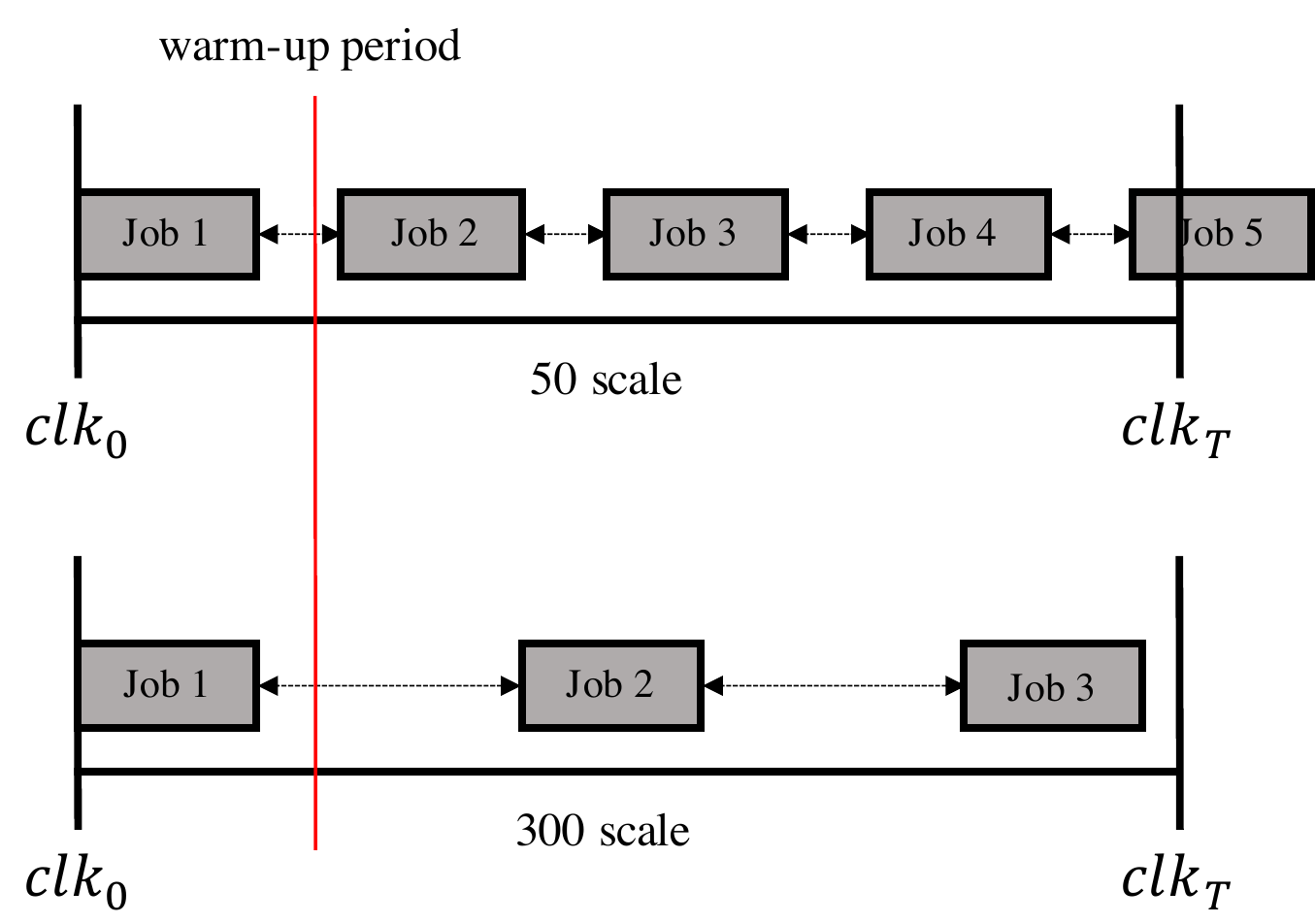}
    \caption{A timeline of multiple jobs injected into the job queue at different scales.}
    \label{fig:jobgen_timeline}
\end{figure}

Figure~\ref{fig:jobgen_timeline} illustrates the timeline that jobs arbitrarily injected into the system based on the scale values. The jobs after the last clock signal are discarded. 

DS3 accounts warm-up period to attain a steady-state. The traces before the warm-up period referring to the initializing phase is neglected. The warm-up period essentially denotes additional duration to reach jobs fully stacked to the job queue. Designing a warm-up period requires domain-expert knowledge. Essentially, the better scheduling policy takes a more extended warm-up period. 

\begin{figure}[!t]
\centering
\includegraphics[height=.27\textwidth]{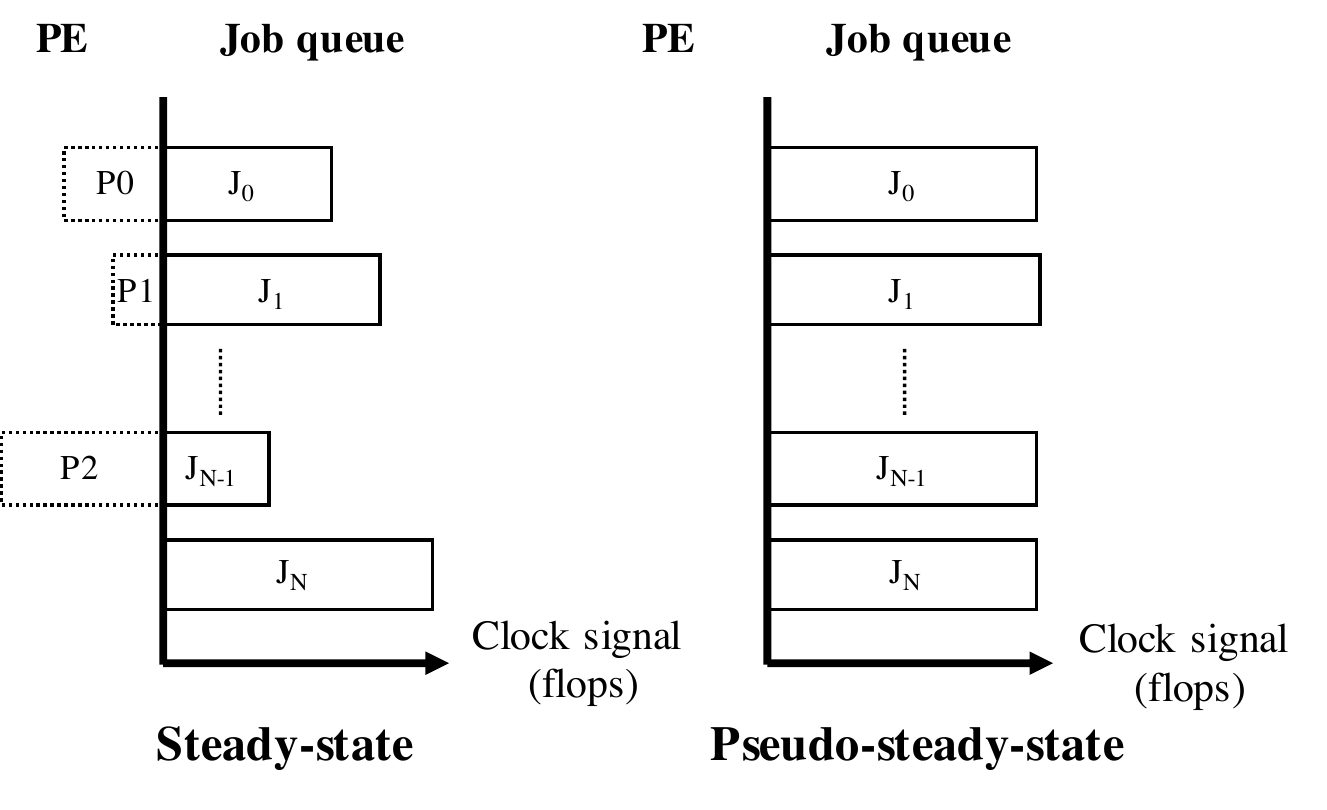}
\caption{A demonstration for steady-state and pseudo-steady-state.}
\label{fig:pss}
\end{figure}

In practice, reaching to initializing phase wastes time to run an episode, particularly in RL training. Sung et al. introduce `pseudo-steady-state' (PSS) that approximates steady-state to reduce the waiting time. As denoted in Figure~\ref{fig:pss}, in PSS mode, the episode starts running from the complete jobs stacked in the job queue. Assume that the system holds up to $N$ jobs, then the PSS starts simulation after generating $N$ jobs in the job queue. In this paper, we evaluate scheduling algorithms with the PSS option.

\subsection{Implementation} \label{sec:proposed:implementation}
This section derives the code snippets of the proposed environment framework. First, we provide the \texttt{GymDS3} class built upon the Gym framework~\cite{1606.01540}.

\lstset{style=mystyle}
\begin{lstlisting}[language=Python]
class GymDS3(gym.Env):
    def __init__(self):
        ...
        
    def reset(self):
        # reset environment

    def step(self, action):
        if ready task list is not empty:
            # select action
        else:
            # select no-action and run simulator
            # update immediate reward
            
        obs = self._get_observation()
        return obs, reward, done, {}
        
    def _get_observation(self):
        ...
        return (job_dags, action_map, env_storage, PEs)
\end{lstlisting}

In the \texttt{step} function, the action can be valid action sets (i.e., task-to-PE mappings) or `no-action' in that the task not in the ready list. For the latter case, the immediate reward is updated. The \texttt{\_get\_observation} function returns environment instances. \texttt{job\_dags} denotes the injected job data, \texttt{action\_map} the take-to-PE mappings, \texttt{env\_storage} the task statuses and performance statistics, \texttt{PEs} the PE information. Next, we demonstrate the actual simulation processing code snippets.

\begin{lstlisting}[language=Python]
# Create a new environment and reset it. 
env = GymDS3(simulation_length, scale)
state = env.reset()

# Create a scheduler
scheduler = get_scheduler(env, scheduler_name)
# If using 'DeepSoCS' as scheduler
if scheduler_name == 'DeepSoCS':
    sess = tf.Session()
    actor_agent = ActorAgent(sess, **kwargs)
    scheduler.set_actor_agent(actor_agent)
	
done = False

while not done:
    action = scheduler.schedule(state)
    state, reward, done, info = env.step(action)
\end{lstlisting}
The environment is generated with total simulation length and scale value. In particularly using neural scheduler (i.e., DeepSoCS~\cite{sung2020deepsocs}), TensorFlow session~\cite{abadi2016tensorflow} and corresponding agent are initialized.

\section{Experiments} \label{sec:experiments}
This section demonstrates the feasibility of gym-ds3 to standard DS3 framework by conducting experiments and two orthogonal directions. (a) We compare average response time for gym-ds3 and DS3 in different job frequencies and provided schedulers. (b) We verify the latency performances in different schedulers with gym-ds3 and DS3 simulations.

\subsection{Experimental Setup}
Throughout the experiments, we use five different types of jobs that modified topology based on the Simple profile from Figure~\ref{fig:top_profile}. The jobs are continuously generated to the exponential distribution controlled by the scale factor. To diminish time for initializing phase, we start evaluation on pseudo-steady-state condition. We set the job queue length to three and simulate for 5,000 clock time (flops). We conduct five trials using different random seeds to produce result data.

The gym-ds3 environment provides heuristic schedulers: Shortest Job First (SJF)~\cite{vasile2015resource}, Minimum Execution Time (MET)~\cite{braun2001comparison}, Earliest Task First (ETF)~\cite{blythe2005task}, and Heterogeneous Earliest Finish Time (HEFT)~\cite{topcuoglu2002performance,bittencourt2010dag,arabnejad2013list}. Also, we include DeepSoCS~\cite{sung2020deepsocs}, which is a hybrid scheduler on Deep RL and heuristic approaches.

\subsection{Average response time} \label{sec:experiments:art}
\begin{figure}[h]
\centering
\includegraphics[height=.28\textwidth]{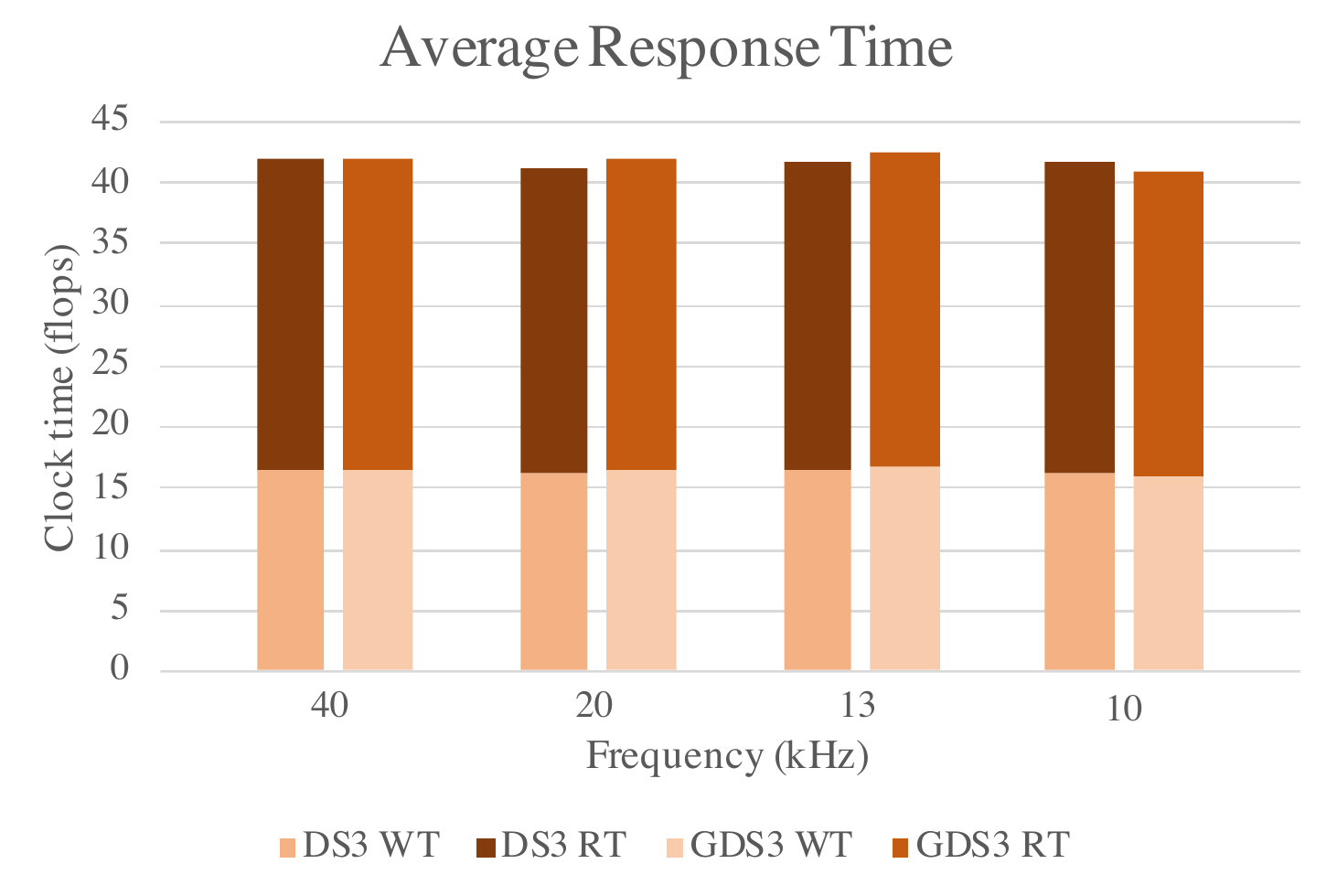}
\caption{Average response time for DS3 and gym-DS3 on different job generating frequencies. Minimum Execution Time scheduler is conducted.}
\label{fig:gymds3_art}
\end{figure}
We extrapolate average response time (ART) using the gym-ds3 environment and standard DS3 framework with different job generation frequencies and schedulers. Figure~\ref{fig:gymds3_art} highlights the task waiting time and task running time using MET scheduler. Considering that the DS3 successfully mimics customer hardware performance, we discover that gym-ds3 simulation demonstrates almost equivalent performance to the DS3 framework. The minor differences can be neglected due to the simulation variances.
\begin{figure}[h]
\centering
\includegraphics[height=.3\textwidth]{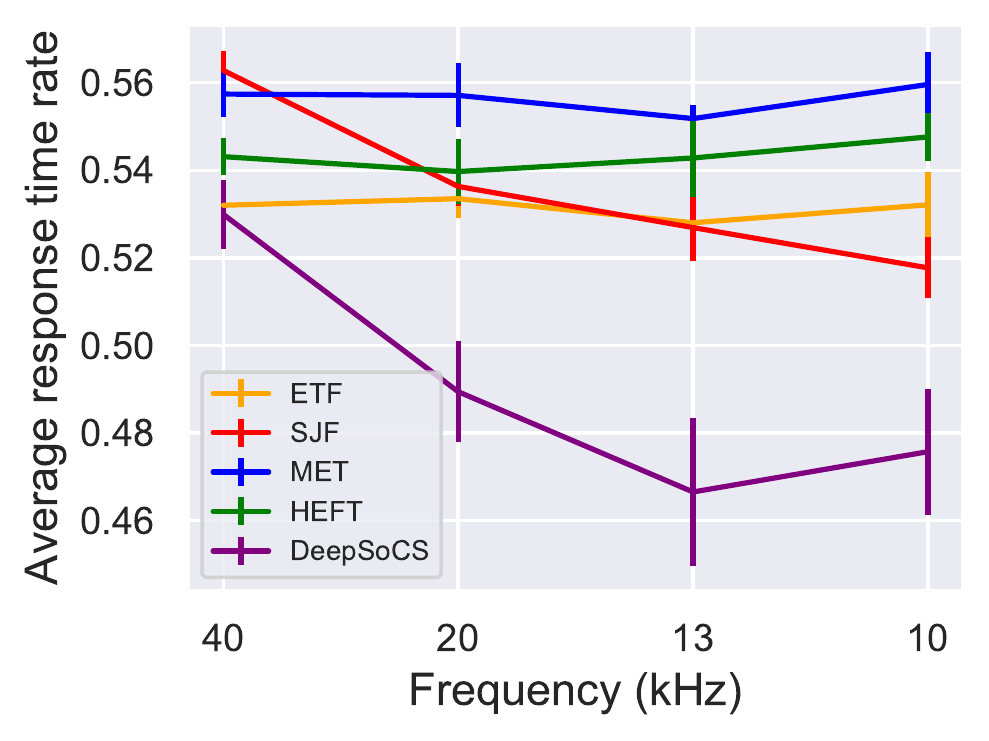}
\caption{A demonstration of average response time on provided heuristic schedulers using gym-ds3 environment.}
\label{fig:gymds3_sched_perfs}
\end{figure}
Next, we evaluate ART on different schedulers and job frequencies using gym-ds3 environment, as illustrated in  Figure~\ref{fig:gymds3_sched_perfs}. The variances are marked with the error bar. Comparing to heuristic schedulers where demonstrate similar ranged performances, DeepSoCS outperforms but has high variances.

\subsection{Performance evaluation}
\label{sec:experiments:perf_eval}
\begin{figure}[h]
\centering
\includegraphics[height=.26\textwidth]{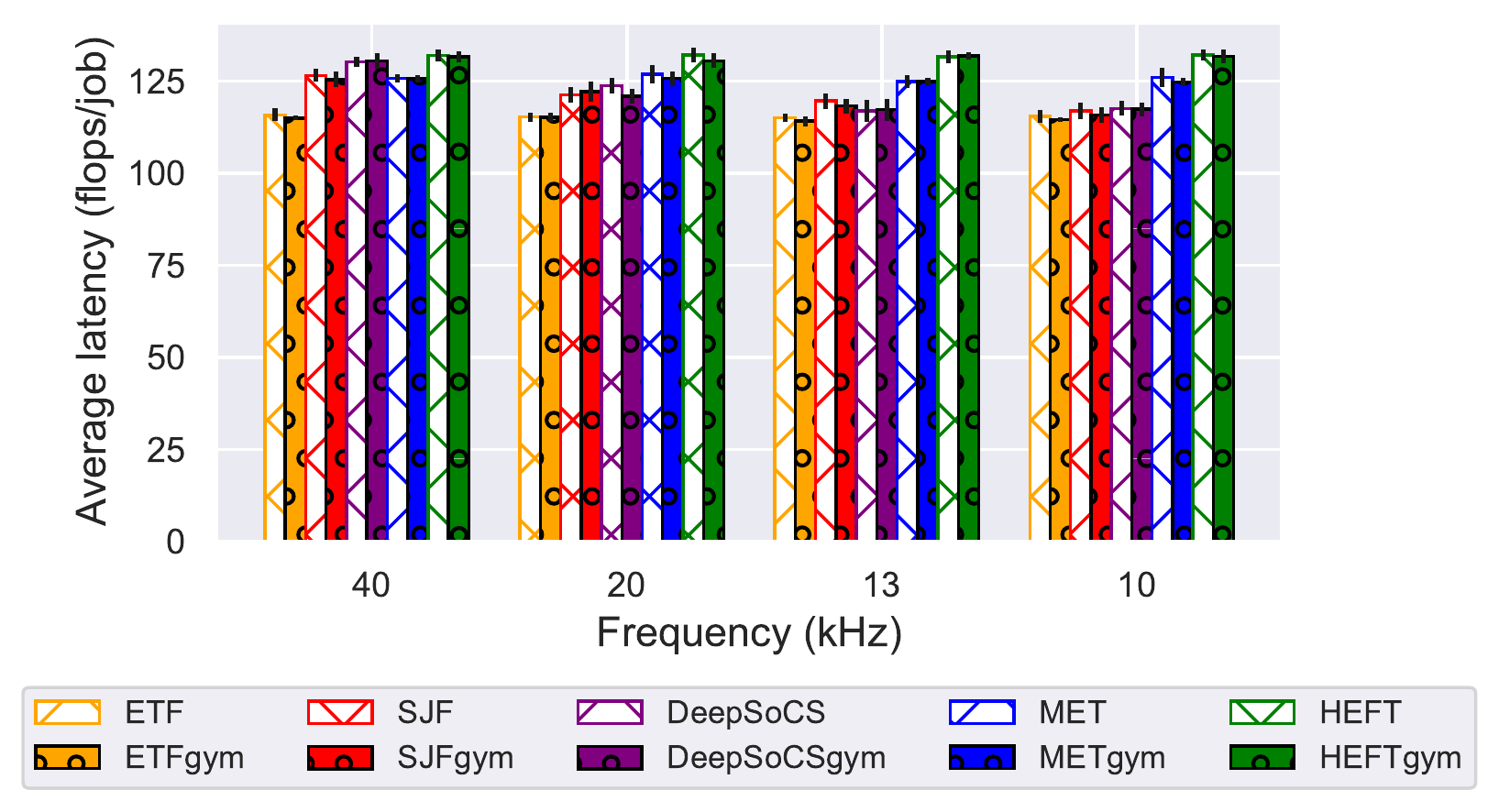}
\caption{A performance evaluation with different schedulers using standard DS3 and gym-ds3.}
\label{fig:ds3_gymds3_eval}
\end{figure}

Subsequently, we compare average latency for different schedulers using DS3 and gym-ds3 environments. Figure~\ref{fig:ds3_gymds3_eval} demonstrates the run-time performances conducted using varied job frequencies.

The run-time performances evaluated on the gym-ds3 environment successfully mimic those on the DS3 framework. Note that both have neglectable differences arisen from the system variances. Upon the experimental results, we can conclude that gym-ds3 validates indistinguishable performances with the standard DS3 framework.

\section{Conclusion}
This paper presents the gym-ds3 environment that provides equivalent functionalities to the DS3 framework. The proposed system operates upon the Gym mechanism, which is comprehensible to RL interaction. We systematically analyze the DS3 simulation and pose challenging standpoints from designing RL agents in DS3 simulation. Furthermore, we experimentally validate run-time performances using various schedulers and job frequencies in gym-ds3 and DS3 and extrapolate almost identical performances.

\section*{Acknowledgment}
The authors would like to thank Hanbum Ko for experiment setup and Jeewoo Kim for valuable discussions.

\bibliographystyle{plain}
\bibliography{main_bib}

\end{document}